# Talking to the brain: Using Large Language Models as Proxies to Model Brain Semantic Representation

Xin Liu[1-5], Ziyue Zhang[1-5], Jingxin Nie[1-5]*

[1]Philosophy and Social Science Laboratory of Reading and Development in Children and Adolescents (South China Normal University), Ministry of EducationCenter for Studies of Psychological Application, South China Normal University; Guangzhou,510631, China.
[2]Center for Studies of Psychological Application, South China Normal University; Guangzhou, 510631,China.
[3]Key Laboratory of Brain, Cognition and Education Sciences (South China Normal University), Ministry of Education.
[4]School of Psychology, South China Normal University; Guangzhou, 510631,China.
[5]Guangdong Key Laboratory of Mental Health and Cognitive Science, South China Normal University; Guangzhou,510631, China.
*Corresponding author Email: niejingxin@gmail.com.

## Abstract

Traditional psychological experiments utilizing naturalistic stimuli face challenges in manual annotation and ecological validity. To address this, we introduce a novel paradigm leveraging multimodal large language models (LLMs) as proxies to extract rich semantic information from naturalistic images through a Visual Question Answering (VQA) strategy for analyzing human visual semantic representation. LLM-derived representations successfully predict established neural activity patterns measured by fMRI (e.g., faces, buildings), validating its feasibility and revealing hierarchical semantic organization across cortical regions. A brain semantic network constructed from LLM-derived representations identifies meaningful clusters reflecting functional and contextual associations. This innovative methodology offers a powerful solution for investigating brain semantic organization with naturalistic stimuli, overcoming limitations of traditional annotation methods and paving the way for more ecologically valid explorations of human cognition.

## 1 Introduction

Psychological experimental paradigms—standardized methods designed to investigate mental processes—rely extensively on carefully crafted procedures and stimuli to test specific hypotheses. First, replicability is often an issue. The dependence on specific stimuli can limit the generalizability of findings, and even minor variations in stimuli may yield different results, casting doubt on the reliability of research outcomes (*1-3*). Second, the experimenter effect (also known as the Rosenthal effect) poses a substantial challenge; Unconscious bias can unintentionally influence participant behavior, twisting results (*4*). Finally, achieving ecological validity—the degree to which findings generalize to real-world settings—is inherently difficult. Laboratory environments often differ markedly from everyday contexts, limiting the applicability of research conclusions (*5*). These limitations highlight the pressing need for innovative approaches in psychological research.

To address these challenges, particularly those related to ecological validity, employing more naturalistic stimuli has emerged as a promising solution (*6-8*). Researches in visual neuroscience has demonstrated the advantages of naturalistic neuroimaging paradigms, which use natural movies or images instead of simple, artificial stimuli (*9-11*). These complex stimuli better reflect real-life scenarios, producing stronger and more reliable brain responses, and providing richer insights into neural representations of objects and scenes (*12,13*). Such paradigms have significantly advanced our understanding of the brain's semantic space. However, they also faces new challenges, especially the requirement for extensive manual annotation of large datasets, which is both time-consuming and resource-intensive.

Recent advancements in artificial intelligence, particularly the development of powerful multimodal large language models (LLMs) like GPT-4 (*14*) and Gemini (*15*), offer a promising solution to this annotation bottleneck (*16*). These models exhibit impressive capabilities across a range of visual and cognitive tasks, often demonstrating human-like performance, including sophisticated understanding and generation of text related to visual content (*17,18*). Crucially, LLMs can automatically process and interpret complex natural stimuli, generating rich semantic descriptions and representations without requiring manual annotation (*19,20*). This automated semantic analysis using LLMs allows researchers to efficiently extract meaningful information from large datasets of natural images or videos, addressing the key challenge in naturalistic paradigms.

To solve these limitations, we propose a novel paradigm that uses artificial neural networks as proxies in psychological experiments to investigate the human visual semantic representation (as shown in **Figure 1**). This paradigm analyzes human brain cognitive function in response to naturalistic stimuli, counting on the capabilities of a multimodal large language model. The data acquisition step occurs before experimental design, where content-rich images or videos serve as the primary stimuli. Next, prompts are designed to guide the LLM in performing Visual Question Answering (VQA) tasks. Based on their advanced image-text processing capabilities (*21-23*), LLMs extracts and encodes image features, generating a semantic representation of each image. Finally, to establish a link between these computational representations and neural activity, generalized linear model (GLM) could be employed to predict blood-oxygen-level-dependent (BOLD) functional Magnetic Resonance Imaging (fMRI) responses for each cortical voxel, using the LLM-derived semantic representation as predictors.

In this study, we initially examined the activation results for the semantic categories "face" and "building", comparing them with previous studies (*24-28*) to validate the feasibility of our paradigm. We then extended our analysis to include dozens of semantic labels to explore stable semantic activation patterns across the cortex. Moreover, we investigated how the semantic hierarchies are presented on the cerebral cortex, showing that both the categories and connotations associated with different semantic levels influence the specific brain regions activated, thereby demonstrating a hierarchical organization of semantic processing in the brain. Finally, be integrating these findings, we constructed a semantic network of the brain and observed distinct activation patterns corresponding to different semantic categories. These findings suggest that artificial neural networks, particularly LLMs, can serve as effective proxies in modeling complex brain functions. By bridging the gap between artificial neural networks and human neural representations, this paradigm offers a promising approach for future psychological research, potentially overcoming many limitations of traditional experimental paradigms.

# 2 Methods

## 2.1 Dataset

This study used data from two distinct sources, involving a total of 11 participants and 78,520 natural image stimuli.

(1) Natural Scenes Dataset (NSD) (*29*): The NSD dataset, collected using high-resolution 7T fMRI, includes data from eight healthy adult participants. These participants viewed approximately 73,000 color natural scene images drawn from the COCO dataset, amount 80 categories. Each participant completed 30 to 40 scanning sessions and viewed roughly 10,000 unique images. The NSD's high-quality image and well-designed experimental protocol provide a robust foundation for our visual stimuli analysis.

(2) Purdue Movie Dataset (*8*): This dataset, obtained from the Integrated Brain Imaging Laboratory at Purdue University, consists of 3T fMRI data from three participants. These participants watched naturalistic movies featuring real-world visual content (e.g., human and animal movement, natural and built environments). Each participant contributed 11.47 hours of fMRI responses to 3.07 hours of movie stimuli. For this study, we extracted individual frames from the movie corresponding to fMRI acquisition time points, resulting in 5,520 natural images.

## 2.2 Multimodal Large Language Model for Semantic Feature Extraction

The Natural Scenes Dataset (NSD) uses stimuli from the Microsoft COCO dataset, which includes 80 easily recognizable object categories (*30*). To extract semantic features from these images, we employed the pre-trained multimodal large language model Blip (Bootstrapping Language-Image Pre-training) (*31*). Its pre-training incorporates graph-text contrastive loss, graph-text matching loss, and language modeling loss, which together enhance its semantic comprehension capabilities, enabling strong performance in image-text retrieval and captioning tasks.

In this study, we used Blip to encode image semantics through Visual Question-Answering approach. For each semantic label we are interested, a specific prompt (e.g., "Is there any face that can be easily recognized in this image?") was designed to query Blip. Blip received both the image and the prompt as input, and its binary "Yes" or "No" response to each prompt was then encoded to represent the semantic presence of the label. This method provides an efficient and targeted approach to extract semantic information aligned with the defined object categories, while also estimating how humans might categorize visual content.

## 2.3 Model fitting

We employed a regression analysis to assess how well the extracted semantic representation could explain participants' brain activity during visual stimulation. For each participant, the semantic representation generated by the large language model (based on image stimuli and corresponding prompts) served as predictors in the regression analyses.

The semantic representation encoding produced binary "yes" or "no" responses for each prompt, which formed our input variables. To address potential bias from imbalanced responses, we balanced the disparities in the number of "yes" and "no" answers. Specifically, we calculated the difference in counts of two answers and then applied random undersampling to remove redundant responses. This process ensured an equal number of two answers instances for each semantic category. Correspondingly, we removed the fMRI data points associated with the discarded semantic responses, aligning the semantic and brain activity datasets. This balancing step ensured an equal number of instances for each semantic category in the regression analysis.

We conducted a GLM regression analysis (*32*) for each voxel on every participant's cortex. This analysis calculated the coefficient of determination ($R^2$) and t-value to quantify the strength and significance of the relationship between each voxel's activity and the semantic features.

After fitting the model at the individual level, we integrated the results at the group level to construct a collective semantic space. Individual regression outcomes were statistically summarized and modeled using a group-level GLM, followed by statistical testing. To control the false positive rate arising from multiple comparisons, we applied cluster correction based on Random Field Theory (RFT) (*33*). Using Monte Carlo simulations (*34*), we determined the appropriate cluster size threshold to correct for multiple comparisons by evaluating the spatial contiguity of activation and considering the spatial correlation of voxels.

For cluster correction, we used a voxel-level significance threshold of $p < 0.05$ with a 3mm full-width at half-maximum (FWHM) smoothing kernel. Based on the Monte Carlo simulations, the minimum cluster volume for significance under a two-tailed test was 77.84 $mm^3$, with a cluster-level significance threshold also set at $p < 0.05$. All data analysis and correction were performed using DPABISurf software (*35*).

## 2.4 Semantic Hierarchy

To construct a semantic hierarchy that reflects human conceptual organization, we used WordNet (*36*), a comprehensive lexical database of English. WordNet organizes words through semantic relationships such as synonymy, hyponymy, and meronymy. We established a category hierarchy based on the hierarchical structure from WordNet. For instance, starting with "human", we extracted its superordinate term "mammal" and subordinate terms "man". Following this approach, we selected a series of hierarchically related labels, resulting in the specific semantic hierarchy: "animal" → "mammal" → "human" → "man".

To examine the neural representation of this semantic hierarchy, we analyzed the brain activation patterns associated with each semantic label. Specifically, we used the extracted image semantic representation to perform GLM regression for each hierarchical label, quantifying the corresponding brain activation signals. We then overlayed the brain activation maps for each label with those of its superordinate label. This superimposition allowed us to visualize overlapping activation patterns across the hierarchical semantic network on the cerebral cortex, revealing the shared and distinct neural substrates of semantically related concepts. This approach provided insights into how the brain encodes semantic hierarchy, offering a deeper understanding of the neural mechanisms underlying the organization and processing semantic information.

## 2.5 Semantic Network Construction

The activation patterns for individual label were projected onto the cerebral cortex to represent a unified spatial semantic network. We then generated a semantic similarity matrix by computing the Pearson correlation between pairwise cortical activations for each label. This similarity matrix, which reflects semantic relationship, was transformed into a distance matrix:

$$distance = 1 - similarity. \qquad (1)$$

This conversion shifted the correlation matrix range from [-1,1] into a distance matrix range of [0,2]. Agglomerative hierarchical clustering (*37*) was applied to the activation results of the 80 semantic labels to explore representational distinctions. This clustering method iteratively merged the most similar clusters by minimizing intra-cluster variance. Finally, we constructed a semantic connectivity network. In this network, nodes represented individual semantic labels, with attributes for name and cluster membership. Edges connected labels that exhibited strong correlations (above a defined threshold of 0.55), indicating potential semantic relationships. This network visually captures the relationships and clustering structure among the semantic labels.

### 2.6 Data availability

The NSD dataset (*29*) is publicly available on https://naturalscenesdataset.org/.Purdue Movie dataset (*8*)is publicly available on https://purr.purdue.edu/publications/2809/1

### 2.7 Code availability

All analyses were performed using Python and DPABISurf (*35*). The fMRI data were analyses with nilearn (https://nilearn.github.io/stable/index.html). Part of model fitting and statistical analysis was conducted using Statsmodels (*38*).

## 3 Results

### 3.1 Validation with Well-Established Semantic Brain Activation Pattern

To validate our approach of using LLMs to model brain semantic representation, we first validated whether our method could replicate well-established findings on brain activation patterns for specific semantic categories. We focused on two prominent categories: faces and buildings, which are known to elicit distinct responses in specific brain regions. Using semantic representations derived from the LLM responses to image stimuli, we predicted brain activation patterns associated with these categories. As shown in **Figure 1**, images containing faces triggered significant activation in the fusiform face area (FFA), a region consistently involved in face processing (*25*). This result confirms that our model accurately captures the established link between face perception and FFA activation. Similarly, images of buildings caused significant activation in the parahippocampal area (PPA), a region involved in processing environmental scenes and buildings (*27*). This finding further supports our model's ability to predict established semantic activations in the brain. Moreover, direct comparison of the activation patterns for "face" and "building" revealed distinct and largely opposing trends. This differentiation demonstrates clear separation between these two semantic categories at the neural level, aligning with previous researches. Our results strongly support the effectiveness of using LLMs as proxies in psychological experiments to model brain semantic processing.

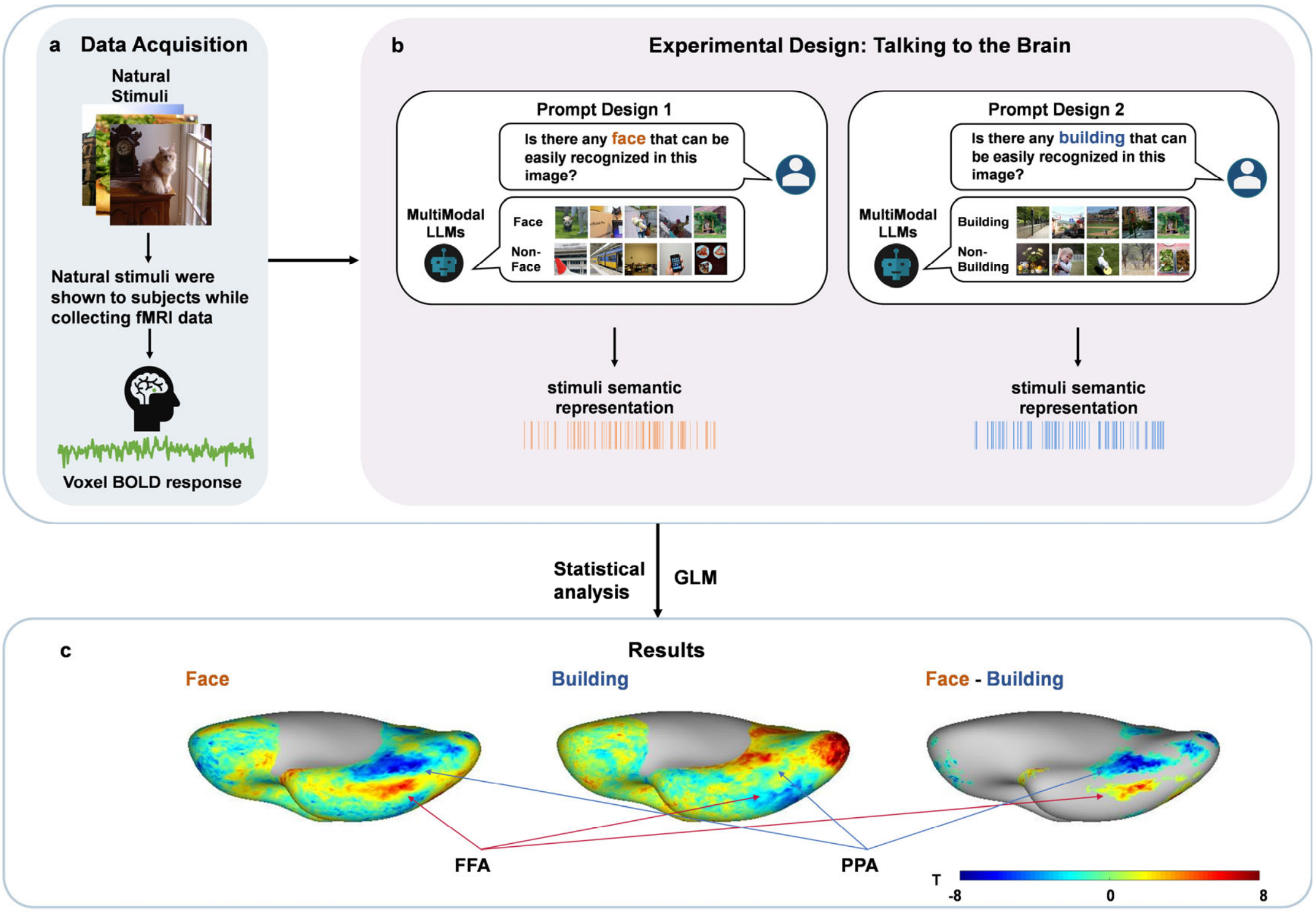

**Figure 1 Using a large language model to extract semantic representation from natural stimuli for fMRI brain activation mapping. a,** Data Acquisition. Subjects viewed natural stimuli for several hours while fMRI BOLD responses were recorded. **b,** Experimental Paradigm. Different prompts were designed based on the tack requirements to query the large language model. The model then extracted the corresponding semantic features of the images and provided binary "yes" or "no" responses, forming the image's semantic representation. A generalized linear model was applied to each cortical voxel, using the semantic representation derived from the large language model to estimate brain activation patterns. The resulting activation maps for each semantic label were generated across the cerebral cortex. **c,** Activation Maps (ventral view, right hemisphere) of "Face", "Building" and "Face-Building", respectively. The color bar represents the t-values derived from voxel-based modeling.

## 3.2 Brain activation patterns of different Semantic labels

Having validated our method's ability to reproduce established semantic activations for faces and buildings, we next explored brain activation patterns across a broader range of semantic categories. We extended our analysis to 80 semantic labels, generating corresponding brain activation profiles (**Supplementary Figure S1**). This expanded analysis revealed distinct activation patterns across various brain regions.

Consistent with the known role of the ventral visual stream in object recognition, we observed widespread activation across most semantic labels within the lateral occipito-temporal cortex. This included the superior temporal sulcus (STS), middle temporal gyrus (MTG), superior temporal gyrus (STG), superior and transverse occipital sulci, anterior occipital sulcus, supramarginal gyrus (SMG), and the intraparietal sulcus (IPS). However, more specific patterns emerged when examining specific semantic categories. For instance, food-related labels (e.g., banana, broccoli) and some tool-related labels (e.g., knife, fork) showed significant activation in frontal regions, specifically the middle frontal gyrus, middle frontal sulcus, and the superior frontal sulcus (SFS). Conversely, sport-related labels (e.g., baseball glove, sports ball) exhibited opposing activation trends in these same frontal regions.

Transportation-related labels also showed significant activation in the superior frontal gyrus. Additionally, we observed a distinction between dynamic and static semantic categories at the temporo-occipital junction (TPOJ): dynamic, outdoor sport-related labels (e.g., sports ball, skateboard) elicited significantly positive activation compared to static semantic labels.

To identify brain regions consistently engaged across a broad range of semantic processing, we superimposed the 80 individual activation maps (**Figure 2**). **Figure 2a** demonstrates that most cortical regions participate in representing at least one semantic category, highlighting the widespread nature of semantic processing. Further analysis revealed stable positive activation during semantic image processing in several key regions, including the early visual cortex (EVC), parieto-occipital sulcus (POS), SMG, fusiform gyrus (FG), STS, PPA, inferior frontal junction (IFJ), and frontal eye field (FEF), as shown in **Figure 2b**. In contrast, consistent negative activation was observed in the IPS, medial superior temporal (MST), and the TPOJ, as shown in **Figure 2c**. These results highlight both shared and distinct brain activation during semantic processing. While regions within the ventral visual pathway and the TPOJ, along with the EVC, showed stable activation across many semantic categories, other regions exhibited more selective responses. This analysis reveals a complex interaction of activation and deactivation across different brain regions during semantic processing, suggesting a sophisticated functional organization beyond simple category-specific responses.

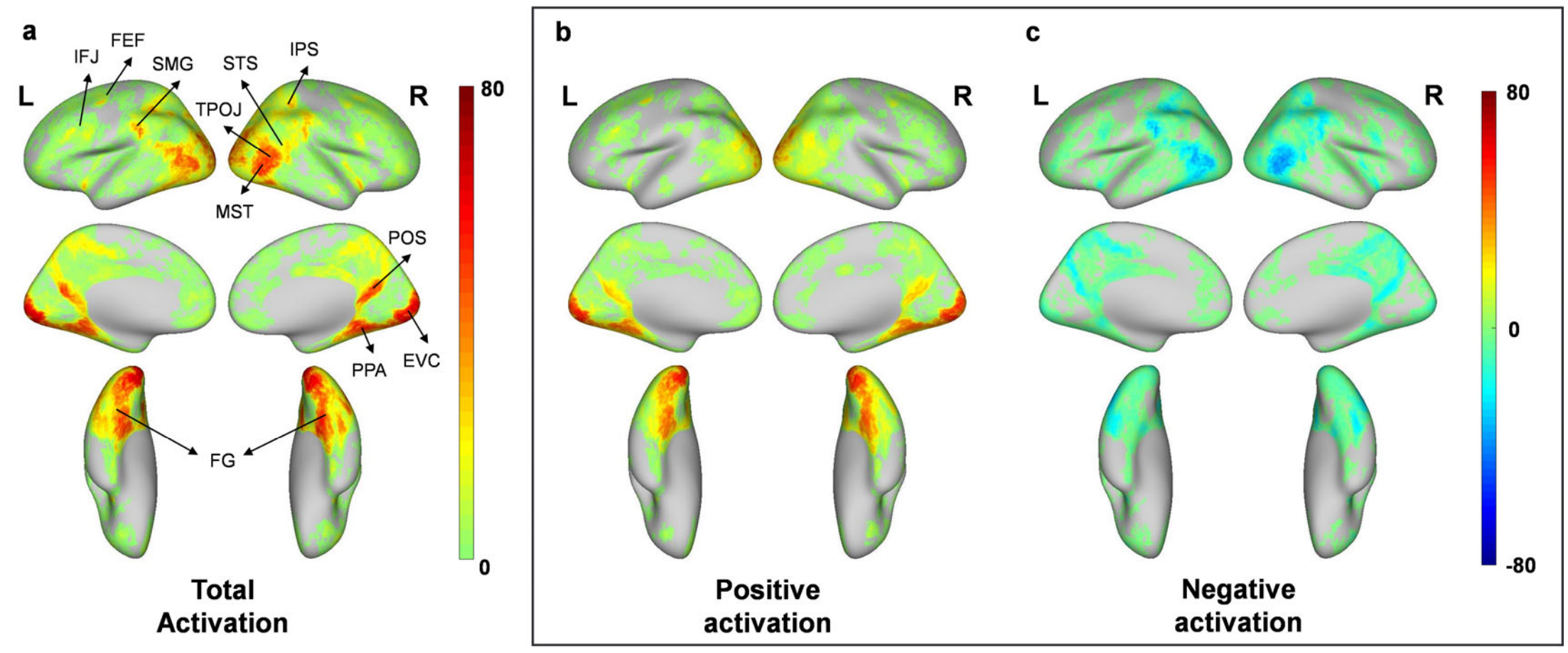

**Figure 2 Cortical activation patterns of 80 semantic labels: overlay of total, positive, and negative activations. a,** Overlayed activation maps of the 80 semantic labels. The color bar indicates the number of semantic labels activated at each voxel. **b,** Overlayed positive activation maps of the 80 semantic labels. **c,** Overlayed negative activation maps of the 80 semantic labels. The color bar indicates the number of semantic labels that are positively or negatively activated at each voxel.

### 3.3 Semantic hierarchical brain activation patterns

Building upon the analysis of individual semantic labels, we investigated how the brain processes hierarchically organized semantic information. We examined the cortical mapping of a four-tiered semantic hierarchy: "animal" → "mammal" → "human" → "man". As shown in **Figure 3**, the activation patterns revealed a clear hierarchical structure: as semantic specificity increased (moving down the hierarchy), additional brain regions with distinct functions were progressively recruited (orange and light blue regions in **Figure 3c**). Each semantic level exhibited unique activation patterns, but significant overlap between levels indicated shared semantic features. Notably, regions such as the TPOJ and the MST were consistently activated across all levels of the hierarchy. This aligns with the whole-brain activation patterns observed in **Figure 2a**, suggesting a general role in semantic processing, regardless of specificity.

We also observed a graded relationship in the overlap of activation patterns between levels: adjacent levels shared more co-activated regions than non-adjacent levels. For example, the overlay in co-activated regions between "animal" and "man" was smaller than the overlap between "human" and "man". Meanwhile, the activation regions revealed by subtracting non-adjacent levels were more extensive than those from subtracting adjacent levels. Specifically, the difference in activation between "mammal" and "human" was smaller than the difference between "animal" and "human". These findings support the idea that higher-level (more general) semantics retain more fundamental features, while lower-level (more specific) semantics require additional neural resources to represent their unique characteristics.

Importantly, the subtraction analyses revealed activation regions corresponding to mutually exclusive concepts within the hierarchy. For example, subtracting "human" activation from "mammal" activation identified regions specifically associated with non-human animals. This was clearly illustrated in the frontal lobe, which showed negative activation for "human" but positive activation for specific non-human mammals such as "bear" and "elephant" (see **Supplementary Figure S1**). This shows the specificity of the activation patterns and demonstrates the ability of our method to isolate neural representations of distinct concepts within a hierarchical structure. Finally, processing more specific semantics engaged not only perceptual regions like the TPOJ and the MST, but also regions associated with the extensional information of these semantics. For example, the STS, a region involved in processing social interaction information (*39*), showed progressively stronger positive activation as we moved from "animal" to "mammal", "human", and finally "man". This pattern reflects the increasing social relevance and complexity of these concepts.

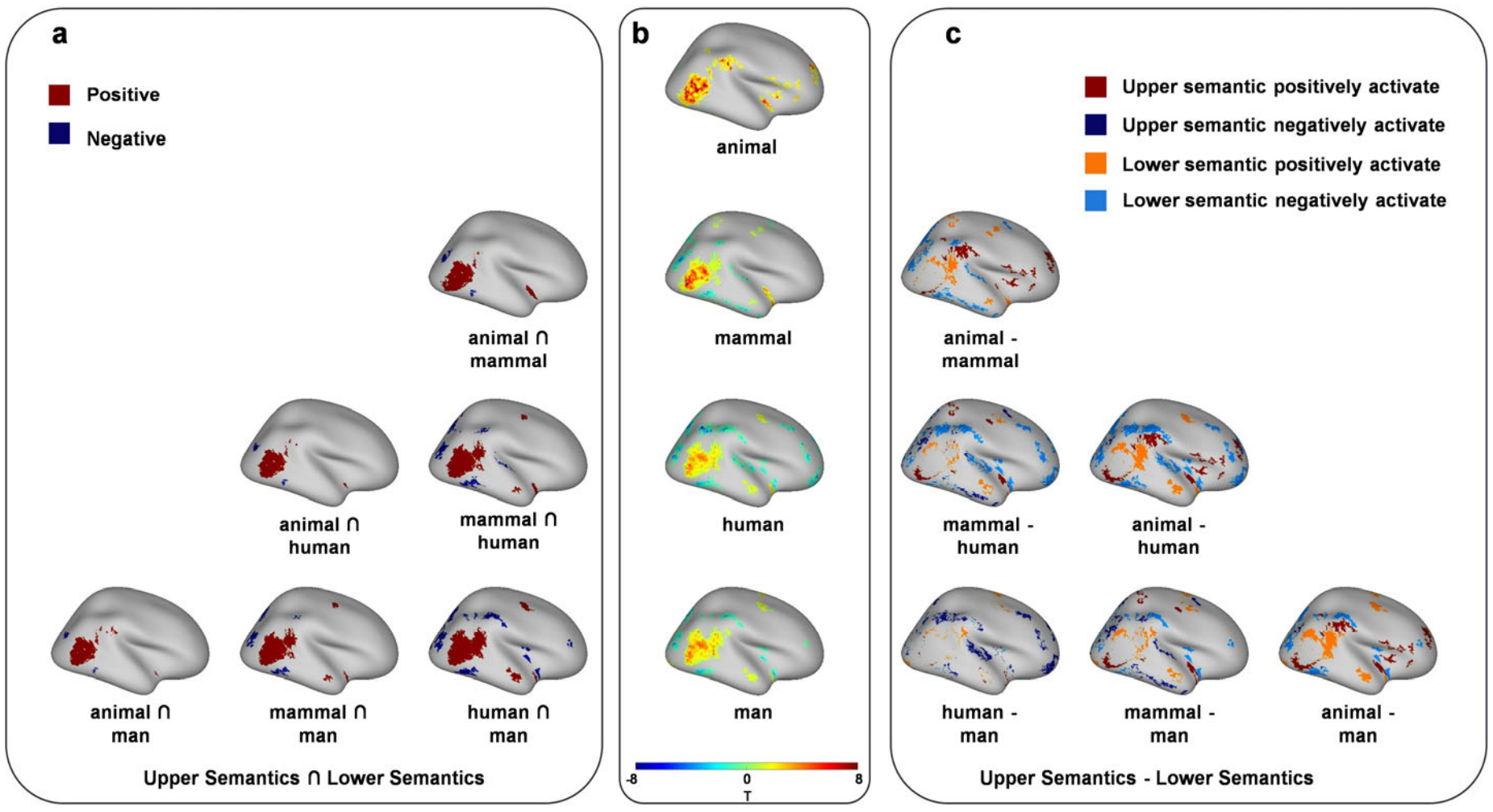

**Figure 3 Cortical activation patterns of upper and lower semantics: overlap and distinct activation zones. a,** The intersection of activation results between upper and lower semantics. Regions where both semantics are positively activated are shown in dark red, while regions where both negatively activated are shown in dark blue. **b,** The activation results of the current semantic level. The color bar represents the t-values derived from voxel-based modeling. **c,** The activation maps of upper semantics versus lower semantics. Regions where only the upper semantics are positively activated are highlighted in dark red, while regions where only the lower semantics are positively activated are shown in orange. Areas where only the upper semantics are negatively activated are shown in dark blue, and areas where only the lower semantics are negatively activated are shown in light blue.

## 3.4 Brain semantic network

To understand how semantic representations are organized within the brain's semantic network, we analyzed the activation profiles of 80 semantic labels across the cerebral cortex. First, we calculated a semantic correlation matrix by computing the Pearson cosine similarity between pairwise cortical activations for each label. Then we applied agglomerative hierarchical clustering to group these labels based on their neural profiles (**Figure 4a**). This analysis identifies five main clusters that reflects distinct functional and contextual associations: (1) Dining table (e.g., cup, banana, pizza), (2) Transportation (e.g., bicycle, bus, traffic light), (3) Indoor scene (e.g., chair, bed, refrigerator), (4) Sports (e.g., frisbee, skis, snowboard), and (5) Others (e.g., backpack, tie, kite). The optimal number of clusters was determined by maximizing the silhouette coefficient.

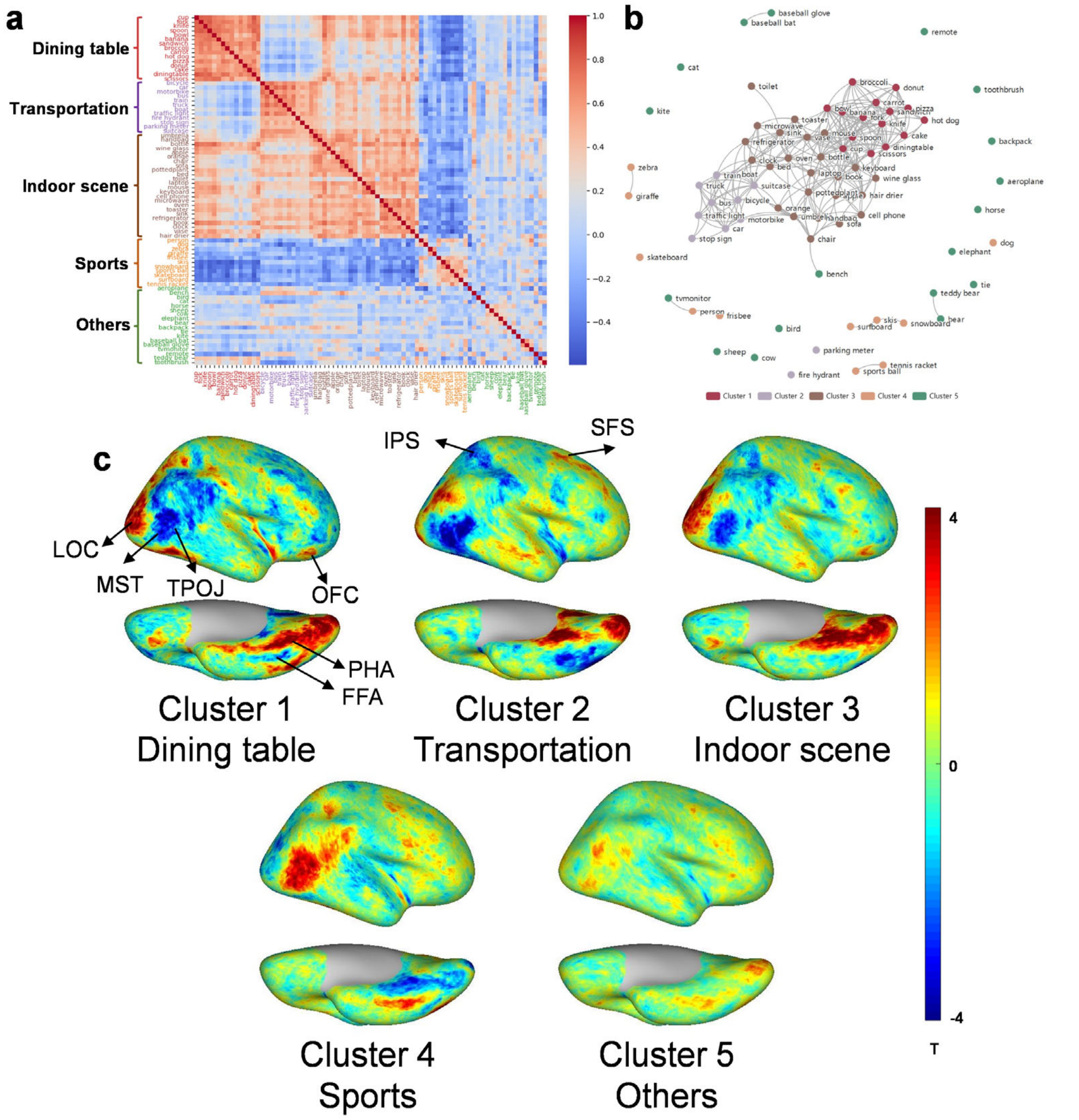

**Figure 4 Brain Semantic Network: Activation Clusters and Correlations. a,** The similarity matrix. The correlation coefficients (color bar) reflect the relationships between activation patterns of the labels. **b,** The semantic connectivity network. The graph visualizing the relationships between labels based on the correlation of their activation patterns. In this graph, nodes represent semantic labels, with different colors indicating distinct clusters, and edges denote the correlations between nodes. **c,** Average whole-brain activation maps (right hemisphere) for each of the five clusters. The color bar represents the t-values derived from voxel-based modeling (left brain see **Figure S2**).

As shown in the semantic correlation matrix in **Figure 4a**, the labels within cluster 1, which mainly related to food and tableware, exhibit highly similar activation patterns across the cortical surface. This cluster shows a closer association with cluster 3, which includes common indoor objects such as furniture, electronics, and household appliances. Notably, items like bottle, wine glass, and apple in cluster 1 show strong correlations with labels in cluster 3. Cluster 2 consists mainly of transportation vehicles and road infrastructure, with strong correlations observed among the vehicles. In contrast, cluster 4, which includes sports equipment, shows high intra-cluster correlation, while the label "person" and others show weaker or even negative correlations with clusters 1, 2, and 3. Cluster 5 contains a more diverse set of labels, including animals, some sports equipment, and household items, however shows limited intra-cluster correlations.

This correlational structure is visualized in the semantic connectivity network shown in **Figure 4b**. Consistent with the clustering analysis, clusters 4 and 5 exhibit relatively low intra- and inter-cluster connectivity, appearing more peripheral in the network. Despite this, connections still form between semantically related items, such as "giraffes" and "zebras", "tennis rackets" and "sports balls". In contrast, clusters 1, 2, and 3 show high intra- and inter-cluster connectivity, forming tightly connected components within the network. Clusters 1 and 3 are semantically closer to each other based on their connection distances within the network.

While these clusters represent distinct semantic categories, their activation profiles revealed both shared and distinct patterns. Clusters 1 (Dining table), 2 (Transportation), 3 (Indoor scene), and 4 (Sports) showed significant activation in the ventral visual stream and TPOJ, suggesting a common role in processing diverse semantic information. However, Cluster 5 (Others) exhibited relatively weaker activation in these regions. More specific differences were observed when comparing activation patterns across clusters. For example, Clusters 1 and 3 showed positive activation in the orbitofrontal cortex (OFC). The TPOJ and the MST showed significant positive activation in Cluster 4 (Sports), while negative activation in Clusters 1, 2, and 3. The lateral occipitotemporal cortex (LOC) showed significant positive activation in Clusters 1 and 3, but a different activation pattern in Cluster 4. The IPS showed significant activation in Clusters 2, 3, and 4. Cluster 2 also exhibited significant positive activation in the SFS. The FG showed positive activation in Clusters 4 and 5, while the PPA showed negative activation in these clusters. Conversely, Clusters 1, 2, and 3 showed strong positive activation in the PPA. The activation patterns of Clusters 1 and 3 were similar, whereas Cluster 2 (Transportation) showed a distinct pattern, particularly with significant negative activation in LOC.

## 4 Discussion

This study introduces a transformative paradigm, "talking to the brain", which effectively bridges the gap between complex real-world stimuli and neural semantic processing. By using the rich knowledge of large language models (LLMs), this novel approach provides a powerful "language" for decoding and modeling the brain's intricate semantic representation. LLMs serve as an advanced intermediary, offering significant advantages compared to traditional methods. Our findings demonstrate the effectiveness of this approach in capturing neural mechanisms associated with specific semantics. The consistency of localized cortical activation patterns with previous research validates the use of LLMs as reliable proxies for modeling neural representations of semantic activation in the brain.

A vital advantage of our LLM-mediated approach is its ability to overcome the inherent limitations of traditional event-related fMRI designs. Unlike traditional methods, which isolate specific stimulus types (*40*), our paradigm is not constrained by pre-defined semantic categories during data acquisition. This flexibility allows us to explore the brain's response to a far broader and more ecologically valid spectrum of concepts, including those not explicitly presented in the experiment. The generalized knowledge of LLMs (*14,15,41-43*) enables investigation about how the brain organizes concepts hierarchically. Our findings suggest that semantic hierarchies those constructed in WordNet, are reflected in the hierarchical organization of the cortex. Importantly, our findings highlight that activation patterns at different hierarchical levels are not always uniform, reflecting the complexity of extensional semantics where higher-level categories don't fully capture lower-level features (*44,45*). For example, activation of "mammal" does not entirely overlay with "human". This may be due to the associations between "human" and other specific object semantics, such as "tools" (*46*).

The hierarchical clustering of the semantic network, derived from our methodology, further demonstrates its capacity to reflect the brain's semantic organization. The resulting clusters not only reflect intuitive groupings of categories (like food and transportation), but also capture more complex, context-dependent associations related to real-world scenarios and activities. The co-clustering of objects from different categories highlights how the brain integrates object functionality with scene recognition, reflecting practical applications of semantic processing. Similarly, the clustering of movement-related terms and animal labels suggests the engagement of motor planning and interaction areas during their processing. This demonstrates the paradigm's ability to simulate brain activity in dynamic contexts. Even seemingly unrelated pairings within clusters suggest the model's sensitivity to delicate, cross-contextual semantic relationships. This associations may arise from shared visual features or life scenarios, further emphasizing the richness of semantic connections captured by our approach.

This innovative approach holds significant promise for advancing neuroscience research. Firstly, it facilitates a more comprehensive and ecologically valid mapping of the brain's semantic space, moving beyond the limitations of artificial stimulus constraints. Secondly, it provides a powerful framework for integrating and analyzing data from diverse neuroimaging experiments that use rich, naturalistic stimuli. By using LLMs to extract consistent semantic representation (*47*), data from multiple studies can be combined, overcoming challenges such as small sample sizes and study-specific designs (*48*), thereby enhancing statistical power and supporting more generalizable conclusions about semantic processing in the brain. This capacity for data integration, based on semantic content rather than low-level stimulus features, represents a significant leap forward in neuroimaging analysis and offers a pathway to more robust and reproducible findings. Moreover, the consistency of our results across different LLM architectures (BLIP and CLIP using VQA and feature similarity method, as demonstrated in **Table S1**) reveals the robustness and generalizability of our method. While we primarily used the VQA approach, the comparable performance with other models shows the potential of various LLM-based implementations for this "talking to the brain" paradigm. The capabilities of advanced LLMs like GPT4 (*14*) and Gemini (*15*) further suggest even greater potential for this methodological direction.

However, it is important to acknowledge the limitations of the current study, including the relatively small sample size and potential bias in data processing. The use of naturalistic stimuli, while beneficial for ecological validity, also introduces complexity. Furthermore, the simple recognition task used during fMRI data collection may not have fully captured the depth of semantic processing. Future research should focus on larger sample sizes and explore more complex task designs to strengthen the robustness and generalizability of these findings.

## Aacknowledgments

We would like to thank the developers of NSD dataset and Puedue Movie dataset as well as developers of the DPABISurf software. This work was supported by the Research Center for Brain Cognition and Human Development, Guangdong, China (No. 2024B0303390003); the Striving for the First-Class, Improving Weak Links and Highlighting Features (SIH) Key Discipline for Psychology in South China Normal University; Key-Area Research and Development Program of Guangdong Province (2019B030335001).

## Aauthor contributions

X.L., J.N. jointly designed the analysis, interpreted the results. X.L., J.N. and Z.Z. wrote the paper. X.L. performed most of the experiments and analyses.

## Onflicts of  interest

The authors declare no competing interests.

## Additional information

Correspondence and requests for materials should be addressed to Jingxin Nie.

Ancillary files for

# Talking to the brain: Using Large Language Models as Proxies to Model Brain Semantic Representation

Xin Liu, Ziyue Zhang, Jingxin Nie*

*Corresponding author Email: niejingxin@gmail.com.

**The file includes:**

Supplementary Text
Figs. S1 to S2
Tables S1

## Supplementary Text

To validate the feasibility of using artificial neural networks as proxies for psychological experiments, we tested two different models to simulate semantically relevant brain activations. Blip (Bootstrapping Language-Image Pre-training) (*31*) was tested using two methods: feature similarity and visual question answering (VQA), while the CLIP was tested through feature similarity alone. CLIP (Contrastive Language-Image Pre-Training) (*42*) is a multimodal model that jointly trains an image encoder (such as ResNet50 or Vision Transformer) and a text encoder (based on the Transformer architecture) to predict the correct match of image and text. CLIP excels in zero-shot learning tasks, achieving accurate predictions without the need for additional training data (*49*). In our experiments, CLIP employed a feature similarity method for image representation encoding. The method inputs crafted prompts and images into a large language model, such as "This image contains a face that can be easily recognized." The text encoder and image encoder then generate two separate eigenvector, respectively. The cosine similarity between these two eigenvectors serves as the semantic representation encoding of the image. Before model fitting, we addressed potential bias from imbalanced responses by balancing the dataset for the VQA method. This step was unnecessary for feature similarity analyses, as they are not affected by response imbalances.

Using the semantic label "face" as an example, we compared the three approaches. Visualization results revealed that both Blip and CLIP achieved high consistency in activation patterns across the cerebral cortex. Notably, all three methods showed significant activation in FFA, a region consistently linked to facial recognition in previous studies, aligning well with our findings. Among the methods, VQA showed strongest performance and was selected for further experiments. Using this approach, we analyzed individual semantic labels. Consistent with previous research, the semantics of "face" and "building" were closely associated with specific brain regions. These results confirm the reliability of our experimental paradigm.

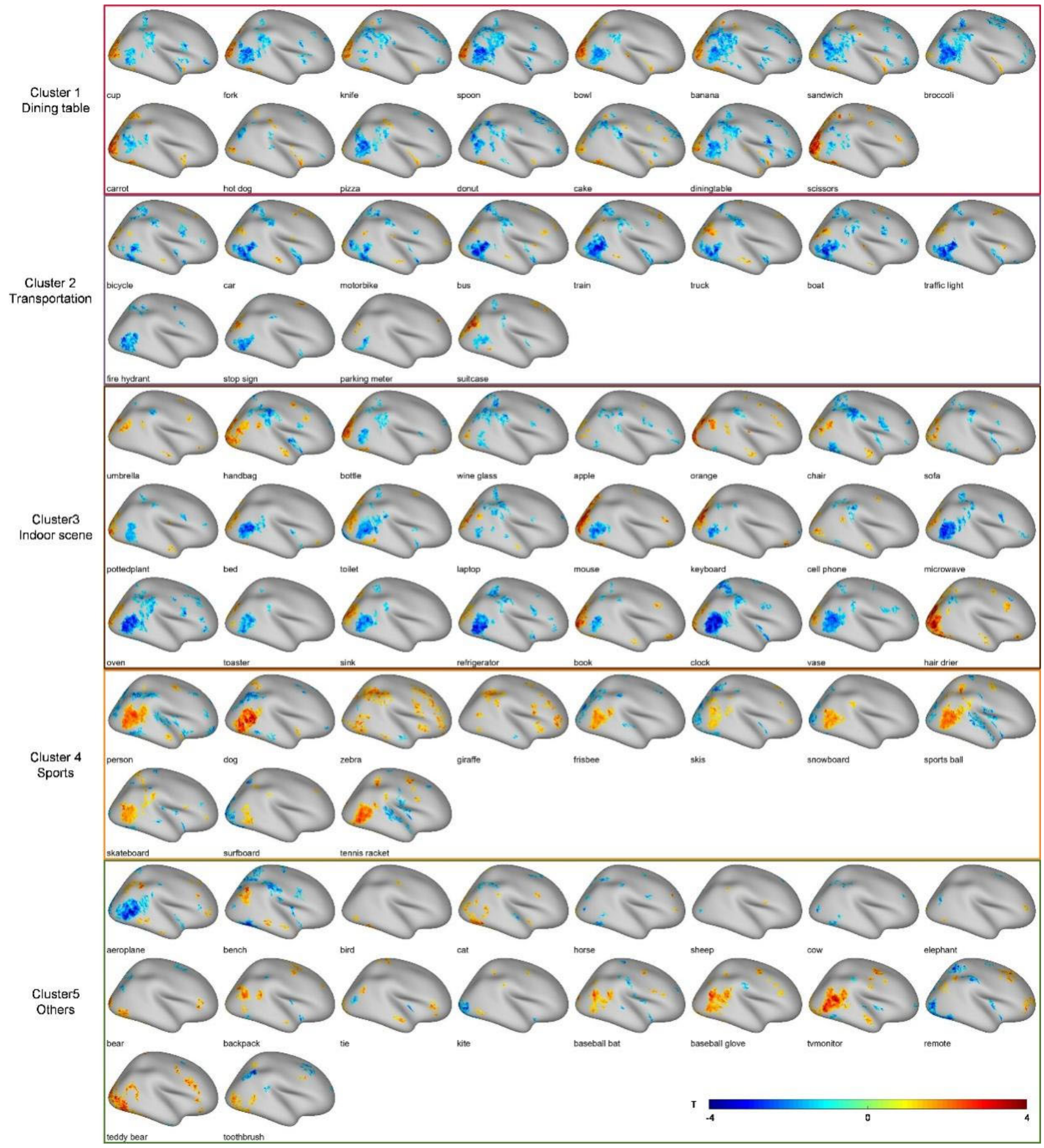

**Figure S1 Brain activation patterns on different semantics**

We calculated cortical activation results for the 80 labels from COCO dataset (*30*), focusing on single semantic categories, The color bar represents the t-values derived from voxel-based modeling. Hierarchical clustering revealed commonalities in semantic activation patterns within each cluster. In cluster 1, which includes kitchen utensils (e.g., fork, knife) and food items (e.g., banana, sandwich), we observed consistent negative activation in the TPOJ region, while positive activation in the LOC region. Cluster 2 comprised semantic labels related to traffic scenarios. These labels showed negative activation in the TPOJ, MST regions at the temporo-occipital junction, and the IPS region in the parietal lobe. Cluster 3 contained semantic labels mainly associated with indoor scenes, including kitchen contexts (e.g., microwave, refrigerator) and electronic products (e.g., mouse, keyboard). These labels showed negative activation in the TPOJ region and positive activation in the LOC region, similar to the activation patterns observed in cluster 1. Cluster 4 included semantics related to animals (e.g., dog, zebra) and sports (e.g., frisbee, skis). Most labels in this cluster showed positive activation in the TPOJ and MST regions. Sports-related semantics and the “person” label displayed similar activation patterns, with significant activation in the STS region. Cluster 5 contained the remaining results. The activation patterns for “aeroplane” in the TPOJ and MST regions resembled those of traffic-related semantics in cluster 2. Meanwhile, the activation patterns for “baseball bat” and “baseball glove” in the TPOJ region were similarly to those of sports-related semantics in cluster 4.

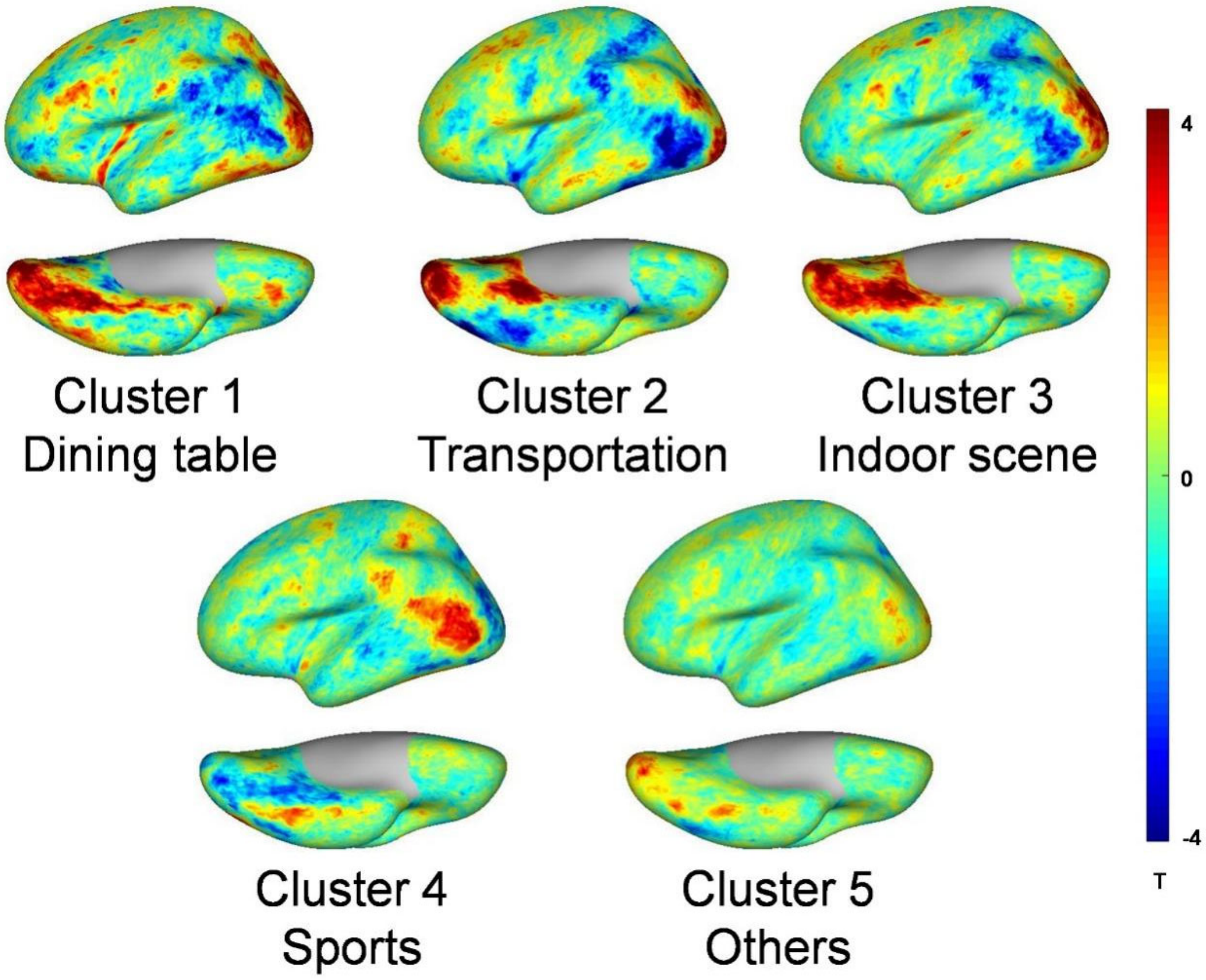

**Figure S2 Average whole-brain activation analysis revealed distinct patterns across the clusters (Left hemisphere)**

The color bar represents the t-values derived from voxel-based modeling. The activation patterns in the left hemisphere are symmetrical to that in the right hemisphere.

**Table S1 Comparison of experimental results across Blip and CLIP using different semantic representation method**

| Model | Method | Mean t-value of FFA |
|---|---|---|
| Blip | VQA | 2.81859 |
| Blip | Feature similarity | 2.79926 |
| CLIP | Feature similarity | 2.09704 |